\documentclass[a4paper,nobind]{ociamthesis}

\correctionstrue

\usepackage[style=numeric-comp, sorting=none, backend=biber, doi=false, isbn=false]{biblatex}
\newcommand*{\bibtitle}{References}

\title{Reinforcement Learning for LLM-based Event Forecasting}
\author{Amit Arnold Levy}
\college{Word Count: 7,828 words}

\degree{Advanced Computer Science}
\degreedate{13th October, 2025}

\begin{document}

\setlength{\textbaselineskip}{22pt plus2pt}

\setlength{\frontmatterbaselineskip}{17pt plus1pt minus1pt}

\setlength{\baselineskip}{\textbaselineskip}

\setcounter{secnumdepth}{2}
\setcounter{tocdepth}{2}

\begin{abstractseparate}
	We use Group Relative Policy Optimization (GRPO), a recently devised sample and memory efficient reinforcement learning method, to finetune pretrained LLMs in the range of 1.5B to 14B parameters equipped with the ability to get current information through the use of a Wikipedia revisions tool, or news summaries, to forecast real events beyond the knowledge cutoff of the LLM, as well as problems made to simulate different aspects of the dynamics of that training.

We use the results of these experiments to comment on the scaling capability of LLMs for forecasting, as well as classify how judgmental forecasting fits into the verifiable/unverifiable domain taxonomy, considering the impact of the inherent aleatoric uncertainty when forecasting future events (e.g. the roll of a die).

As a result of the GRPO training, we manage to bring a 1.5B parameter transformer (Qwen 2.5 1.5B) to forecasting performance superior to Claude Sonnet 3.5 over the same dataset as measured by cross entropy from the market agreed probabilities. We also discuss various dead ends on the path to this result.
\end{abstractseparate}

\begin{romanpages}

\maketitle

\flushbottom

\tableofcontents

\end{romanpages}

\flushbottom

\chapter{\label{ch:1-intro}Introduction} 
DeepSeek R1 \cite{deepseekai2025deepseekr1incentivizingreasoningcapability} used Reinforcement Learning to solve math and coding problems at the level of human experts, achieving a significant breakthrough in ML.

Work in a similar vein inside labs such as OpenAI and GDM has achieved performance at the level of the very best human experts in those same domains \cite{Alex_Wilkins,olympiad1}. For a long time AGI has been the ultimate goal of the field of AI \cite{turing2007computing}, but it's not clear how even superhuman AI in the fields of easily verifiable coding and mathematics problems could be used to bootstrap to general artificial intelligence. %

Forecasting is a domain that is more general than the perfectly verifiable domains of math and coding problems yet still in a gray zone of verifiability - after all, while inherently forecasting questions may not have a ground truth at the moment of prediction, they will eventually have a ground truth by their nature. As such, they can eventually be verified.

\section{Creating Ground Truths for Forecasting}
There are two clear options for the ground truths of forecasting questions, such as "If I roll a six-sided die, will I roll a 6?". The ground truth can either be the result of the event - 1 (Yes) or 0 (No), in this case with 1/6 odds of a 1, or 5/6 odds of a 0. Alternatively it can be the ground truth probability of the event that the LLM should predict, in this case $p=1/6$.

Of course, the second option - the ground truth probability of the event happening, which is the aleatoric uncertainty of a Yes resolution, is in most cases unknowable, and so has to be approximated. On the other hand, simply providing the ground truth of a single event may punish the LLM even when it provides a correct probability. For instance, while predicting 1/6, under a proper scoring rule, would give the LLM the most expected reward if it indeed predicts 1/6, in practice a single training step may reward the LLM the most for predicting 1.0 and punish it for a 1/6 prediction. And regardless of luck, the maximum reward will never be given for correctly predicting 1/6.

To test which of these approaches works better, we have a separate set of experiments working in a simulated forecasting domain in which we have full flexibility to test both options and compare their tradeoffs.

\section{Providing Context to LLMs}
LLMs without added context are naturally, at least until the problem of Continual Learning / Incremental Learning \cite{Wang_Zhang_Su_Zhu_2024} is solved, limited to the data available in their training sets. This knowledge is not sufficient to provide accurate forecasts for most questions, and human forecasters often hunt down information they feel is relevant for their forecasts. For instance, a human forecaster trying to predict the outcome of an election may look at polling data, economic data, statements by the candidates, et cetera.

In this work we examine two different methods to provide information relevant to the LLM. The first is to use a different LLM with internet access - specifically different models from Perplexity, Sonar and Sonar Reasoning Pro, to summarize relevant information available online that is question relevant for the LLM being trained. 

The second method is to provide the LLM with the ability to tool call a "Wikipedia Revisions" tool, which allows it to read a Wikipedia article of its choice at a certain timestamp, as long as it is not after the timestamp at which the prediction is being simulated to have taken place. This prevents temporal leakage and as such allows using questions that have already been resolved, but is significantly more expensive and slower to train, and inserts other issues into training (e.g. Wikipedia's API may be down, rate limits, etc).

\section{Training LLMs to Forecast using GRPO}
The primary and most successful pipeline we arrive at to train an LLM to perform judgmental forecasts is to take a dataset of the 512 highest volume binary questions on the prediction market website Polymarket, summarize information for each question using Sonar Reasoning Pro from Perplexity, and take the current market price as the ground truth, assuming it approximates the true aleatoric probability of the event given information available online \cite{Fama_1970}. We then use GRPO \cite{shao2024deepseekmathpushinglimitsmathematical} with Qwen 2.5 1.5B - 7B \cite{qwen2025qwen25technicalreport} as the underlying LLM and Negative Cross Entropy as the reward. We achieve under a limited context constraint (a few paragraphs of summarized information) for both our finetuned Qwen 2.5 models and the comparison model strong results. Specifically, our small RL finetuned models significantly beat the larger Sonnet 3.5 model, even though it has a later knowledge cutoff, which has been shown to be an advantage in LLM forecasting \cite{halawi2024approaching}.

\chapter{Background}
\label{ch:2-background}

As background, we will introduce the field of judgmental forecasting as a problem domain, describe the difficulties with benchmarking it and provide an overview of the specific reinforcement learning algorithm we use, Group Relative Policy Optimization (GRPO).

\section{Judgmental Forecasting}
Forecasting is the problem of predicting the future, for example, who will win a football game, who will win the 2028 US presidency or when the first man will land on Mars.

It can be separated into two broad types, time series forecasting and judgmental forecasting. We focus on judgmental forecasting, as covered by \cite{tetlock_superforecasting_2015}. In its most common form, and the form we focus on, it involves a binary question (True/False), such as "Will Donald Trump win the 2024 US presidential election?" that the answer for which will be known conclusively in the future but isn't known at the time the question is being predicted.

\section{Proper Scoring Rules}
In answering the question the forecaster provides a probability for the question to resolve to True/Yes/1. Once there are many predictions collected from a forecaster, and sufficient time has passed for the events in question to resolve, their performance can be evaluated in different ways, most commonly a Brier score \cite{brier_verification_1950}, which is the mean of $(o_i-p_i)^2$ where $p_i$ is the forecaster's prediction for question $i$ and $o_i$ is the true outcome, which is $0$ or $1$. An unskilled predictor who always predicts 50\% odds for any event will achieve a brier score of $0.25$, and a lower score is better.

As shown by \cite{tetlock_superforecasting_2015} and \cite{pm_acc_2003}, human forecasts, both independently and especially when aggregated (naively or through market mechanisms) are surprisingly accurate, showing that this domain is tractable. Furthermore there is variance between different people and the average accuracy of their predictions (as measured by e.g. their Brier scores).

Brier Score, as well as Cross Entropy \ref{eq:cross_entropy} and Kullback–Leibler divergence (KL divergence) \ref{eq:kl_divergence} are all known to be proper scoring rules, which means that someone attempting to maximize their reward / minimize their loss, when their predictions are being graded through one of these scoring rules, is incentivized to provide their true belief in the probability of the outcome being predicted.

\begin{equation}
H(P,Q)
=
-\sum_{x \in \mathcal{X}} P(x)\log Q(x)
\label{eq:cross_entropy}
\end{equation}

\begin{equation}
D_{\mathrm{KL}}(P \parallel Q)
=
\sum_{x \in \mathcal{X}}
P(x)\log\left(\frac{P(x)}{Q(x)}\right)
\label{eq:kl_divergence}
\end{equation}

\section{Epistemic and Aleatoric Uncertainty}
Uncertainty in forecasts, and in probabilities themselves when using the Bayesian framing, can be separated into epistemic uncertainty and aleatoric uncertainty. Epistemic uncertainty is uncertainty as a result of things one does not know, but could know in principle, for instance, how many sides a die has. Aleatoric uncertainty on the other hand is irreducible and results from randomness inherent to the problem - for instance, which number a fair 6 sided die will land on (assuming this is beyond our ability to deterministically model).

\section{Reinforcement Learning for LLMs}
Reinforcement Learning is the subset of Machine Learning characterized by delayed, intermittent rewards and trial and error based learning \cite{sutton}. When applied to LLMs that produce tokens, the former means that instead of having a reward/loss after each token, it is sparse and so received every potentially thousands of tokens, and it is not known how each individual token contributed to this reward.

While Transformer pretraining (which is supervised) has been known to be effective since \cite{DBLP:journals/corr/VaswaniSPUJGKP17}, Transformer reinforcement learning only had its first large documented successes with the \cite{shao2024deepseekmathpushinglimitsmathematical} and \cite{deepseekai2025deepseekr1incentivizingreasoningcapability} papers by DeepSeek, in the verifiable domains of easily verifiable math and coding problems.

One of the innovations leading to the breakthrough was GRPO, a method in which many different solutions for a single problem are generated by an LLM (which is stochastic). The different solutions are graded using an automatic grader, for instance by comparing the final answer (ignoring "reasoning" tokens) to a ground truth. Then the different completions are weighed as positive or negative finetuning examples depending on the reward each achieves as compared to the other completions for this problem, see Equation~\ref{eq_adv}. As such, in a hard question, if the LLM generated 31 incorrect solutions and one correct solution, the reasoning trace of the one correct solution will be strongly strengthened and the LLM will be more likely to output reasoning traces like it in the future, and less likely to output the reasoning traces leading to incorrect solutions.

\begin{equation}
\hat{A}_{i,t}
=
\tilde{r}_i
=
\frac{r_i-\operatorname{mean}(\mathbf{r})}
     {\operatorname{std}(\mathbf{r})}
\label{eq_adv}
\end{equation}

\begin{figure}
    \centering
    \includegraphics[width=1\linewidth]{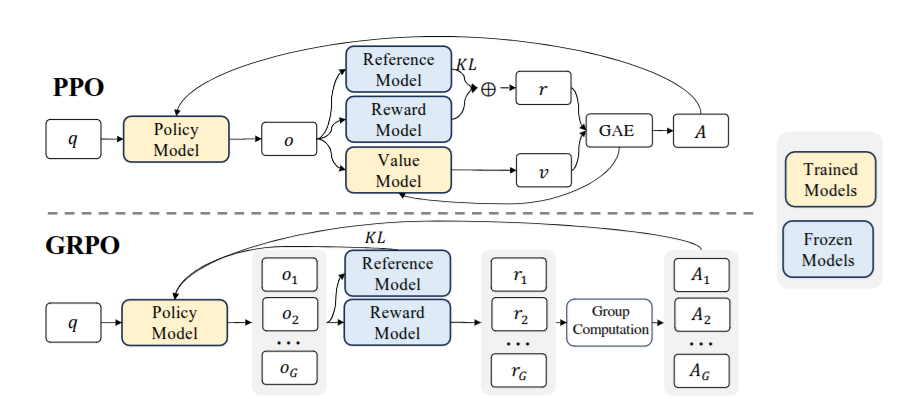}
    \caption{Figure illustrating GRPO in comparison to the more commonly known PPO, taken from \cite{shao2024deepseekmathpushinglimitsmathematical}.}
\end{figure}

An advantage of GRPO over previous methods is that it doesn't require a "value model" - another instance of the LLM kept in memory and used to compute a baseline reward for each problem (since in GRPO the baseline is computed from the same generations used for training), saving about $40\%$ of the memory cost.

\section{Knowledge Cutoffs and Forecasting Benchmarks}
The typical way forecasting ability is measured for humans is that predictions are collected before the events have "resolved", and then the accuracy metrics are only computed after resolution, which can be months or years later. As such there is no worry of "cheating" - nobody knows the answer yet. The disadvantage of this method is that it is very slow since for long term forecasting questions (e.g. US election) we would have to wait years to see if the predictions were accurate.

One way this can be addressed for LLMs is to take advantage of the "knowledge cutoffs" LLMs have \cite{cheng2024dateddatatracingknowledge}. The organizations that train LLMs typically announce when the data used for the dataset was collected, and as a result the LLMs cannot have knowledge of events past that date through their pretraining. As such, one could give an LLM forecasting problems for questions  that have already happened and as such can be immediately evaluated, without worrying they will already have knowledge of how they truly resolved. Of course, in such a scenario the LLMs cannot be given some tools, such as free internet access, or the possibility of leakage becomes possible again.

\section{Prediction Markets}
Prediction markets such as Polymarket, Kalshi and Manifold \cite{hanson_combinatorial_2003, arrow_promise_2008}, allow many forecasters to aggregate their predictions together using market mechanisms. In the typical implementation for a binary question, the market price is between 0 and 100 cents. If it is X cents, that implies an $X\%$ probability for the event to resolve yes. Participants are incentivised to buy the price up if they believe the probability is higher, or down if it is lower. This is because if the event resolves Yes, anyone who has bought a "Yes" share gets 100 cents per share, and anyone who bought a "No" share gets 0 cents. The inverse happens in the case of a No resolution.

As would be implied by the Efficient Market Hypothesis (EMH) \cite{Fama_1970}, prediction markets are currently the most accurate method we have for calibrated probability forecasts, with the alternatives being e.g. individual expert predictions or polls \cite{BERG2008285}. As such for this thesis we consider them the best method to approximate the unknown ground truth probabilities, in the cases where we avoid using the true resolutions.

\chapter{\label{ch:2-related_work}Related Work}

\section{Reinforcement Learning for LLMs}
Recently there have been significant advancements in the success of using reinforcement learning to improve the capability of LLMs in verifiable domains, such as mathematics problems \cite{shao2024deepseekmathpushinglimitsmathematical} and coding problems \cite{deepseekai2025deepseekr1incentivizingreasoningcapability}. Verifiable domains are domains in which it is possible to efficiently calculate a reward for an LLM generated solution, for instance using an automated proof checker (e.g. Lean) in the case of mathematics, or running the code on test cases in the case of coding.

A specific advance is Group Relative Policy Optimization (GRPO) \cite{shao2024deepseekmathpushinglimitsmathematical} which requires only about 60\% of the memory requirements of PPO, because it does not require a critic. There has also been open source work on adapting GRPO for environments when the LLM has access to tool calls \cite{brown_verifiers_2025} which we make use of.

There are also indications of significant other advancements along similar lines, but they are proprietary to labs such as Google DeepMind, Anthropic and OpenAI and the details are not publicly known.

\section{Forecasting using LLMs}
There has been a significant amount of work on using LLMs for forecasting, including judgmental binary forecasting as we focus on here. Recent advancement has been focused on using more modern LLMs, with more recent forecasting questions to match (to avoid temporal leakage), and on increasingly elaborate Information Retrieval (IR) pipelines, see \cite{yan2024autocastenhancingworldevent} \cite{halawi2024approaching}. Notably \cite{halawi2024approaching} has achieved results approaching the aggregate performance of human forecasters when restricting to questions that are easier for the model and harder for humans, such as highly uncertain questions closer to the date of the LLMs knowledge cutoff.

There have also been attempts contemporaneous with ours to use reinforcement learning, including GRPO, to train smaller LLMs to forecast, but results have been weaker, not beating the results achievable by larger pretrained LLMs \cite{turtel2025outcomebasedreinforcementlearningpredict}. We believe that is primarily because they tried to use the final resolution as the ground truth (something that is affected by aleatoric noise). As we show, better learning results are achieved by using high volume prediction market prices as the ground truth.

\section{Benchmarking Forecasting Accuracy}
There have been different attempts to create static, or automatically updating datasets of forecasting questions and their resolutions as a way to benchmark LLM accuracy. These include \cite{zou2022forecastingfutureworldevents} and \cite{lu2025evaluatingllmsrealworldforecasting} among many others.

Since a constantly changing benchmark has obvious weaknesses, and a naively constructed static dataset of forecasting questions quickly becomes irrelevant as LLMs are trained with later knowledge cutoffs, there have also been attempts to create creative benchmarks without this problem. One example is \cite{paleka2025consistencycheckslanguagemodel} which instead of checking the true resolutions of already resolved forecasting questions instead checks the self consistency of different predictions of an LLM, which is something that can be checked even on questions that have not yet resolved. Methods such as these have their own disadvantages, for instance they can be gamed in different ways, as is addressed by \cite{paleka2025consistencycheckslanguagemodel}.

\chapter{\label{ch:3-methodology}Methodology} 

\section{Testing If Learning is Possible Despite Uncertainty}
In preparation for training LLMs to perform judgmental binary forecasting, we need to acknowledge that part of the uncertainty in forecasting the future is aleatoric and as such irreducible. The ideal would be for the model to predict the aleatoric uncertainty. For instance, if the question is, will a fair 6-sided die roll a 6? While there is no better answer than $p\approx0.167$, the ground truth will be either 0 or 1. We need to see if in this environment it is still possible for the RL reward to improve the LLMs capability.

To test this, we will start with the GSM8K dataset (Grade School Math dataset) \cite{cobbe2021gsm8k}, a dataset of simple mathematics word problems, which we further filter down to be restricted to answers in the range 25 to 75. We will then train Qwen 2.5 1.5B in three modes:
\begin{enumerate}
    \item Regular: The model is rewarded for both providing an answer in the correct format, and correctness. This is meant to simulate the case with no aleatoric uncertainty, or in other words, when we have the aleatoric probability itself as the ground truth to be predicted.
    \item Randomized: The answer of each question in the range 25 to 75 will be treated as a probability of a Bernoulli distribution (0.25 to 0.75), and the ground truth will be 100 or 0. This is meant to simulate the forecasting case - we do not know what the correct probability to predict is, only what ended up happening in the single timeline we have access to.
    \item Format Only: We don't give the model any reward beyond encouraging it to give an answer in the correct format. This is so we can compare to the case when we have no useful training data at all.
\end{enumerate}

What we want to see is where is the performance of Randomized in the range between Regular and Format Only.

\section{Real Event Datasets}

\subsection{Unresolved Events}
In order to train and evaluate forecasting on events not yet resolved, we use Polymarket market prices for high liquidity markets as the best approximation we have for the true ground truth probability. We use Polymarket's Gamma Markets API to get the 512 highest volume markets and in all cases train through them with a single epoch. In the primary experiment we add a short summary of additional information to each question generated by Perplexity's Sonar Reasoning Pro, which forms the low information setting. In the high information setting we use both Sonar Reasoning Pro and the base Sonar model to generate two different, longer summaries and concatenate them. The high information regime results are available in the appendix.

\subsection{Resolved Events}
For resolved events we take historical questions from Manifold and Metaculus that were created after the knowledge cutoff of our models and have already resolved. To provide additional information to the model, we give it access to the Wikipedia Revisions tool described below. This was in practice the first experiment performed, but for reasons elucidated by the aleatoric uncertainty experiment, was not as effective as the unresolved events setting, and as such will be described in the appendix.

\section{GRPO Specifics}
To train the models, as mentioned, we use GRPO, which is described in Section~\ref{ch:2-background}. Specifically we perform full fine tuning using GRPO over the datasets above, with a format reward function, as well as a negative quadratic error or negative cross entropy depending on the experiment as the rewards.

For information on tool calling (Wikipedia Revisions), hyperparameters and compute, see Appendix~\ref{section:experiment_details}.

\chapter{\label{ch:4-results}Results} 

\section{Effect of Aleatoric Uncertainty on Training}
A crucial difficulty of forecasting compared to other simpler verifiable domains such as code or math, is that there is randomness in the answer from aleatoric uncertainty. If the question to be forecasted is, "Will the fair coin come up heads?" the probability that the model should predict is 50\%, but the ground truth answer will be 0.0 or 1.0 and so the reward will be accordingly random. In fact, any generation by the model different from 50 may get rewarded more, if it happens to luck out and be on the correct side.

We want to test how much such randomness affects the capability of the model to learn a task through RL. As such we take the GSM8K dataset, but introduce a new mode in which the answer, which is initially between 0 and 100, is treated instead as a probability and "collapsed" to either 0 or 100 (representing a No or Yes resolution) with respect to that probability.

\begin{figure}
    \centering
    \includegraphics[width=1\linewidth]{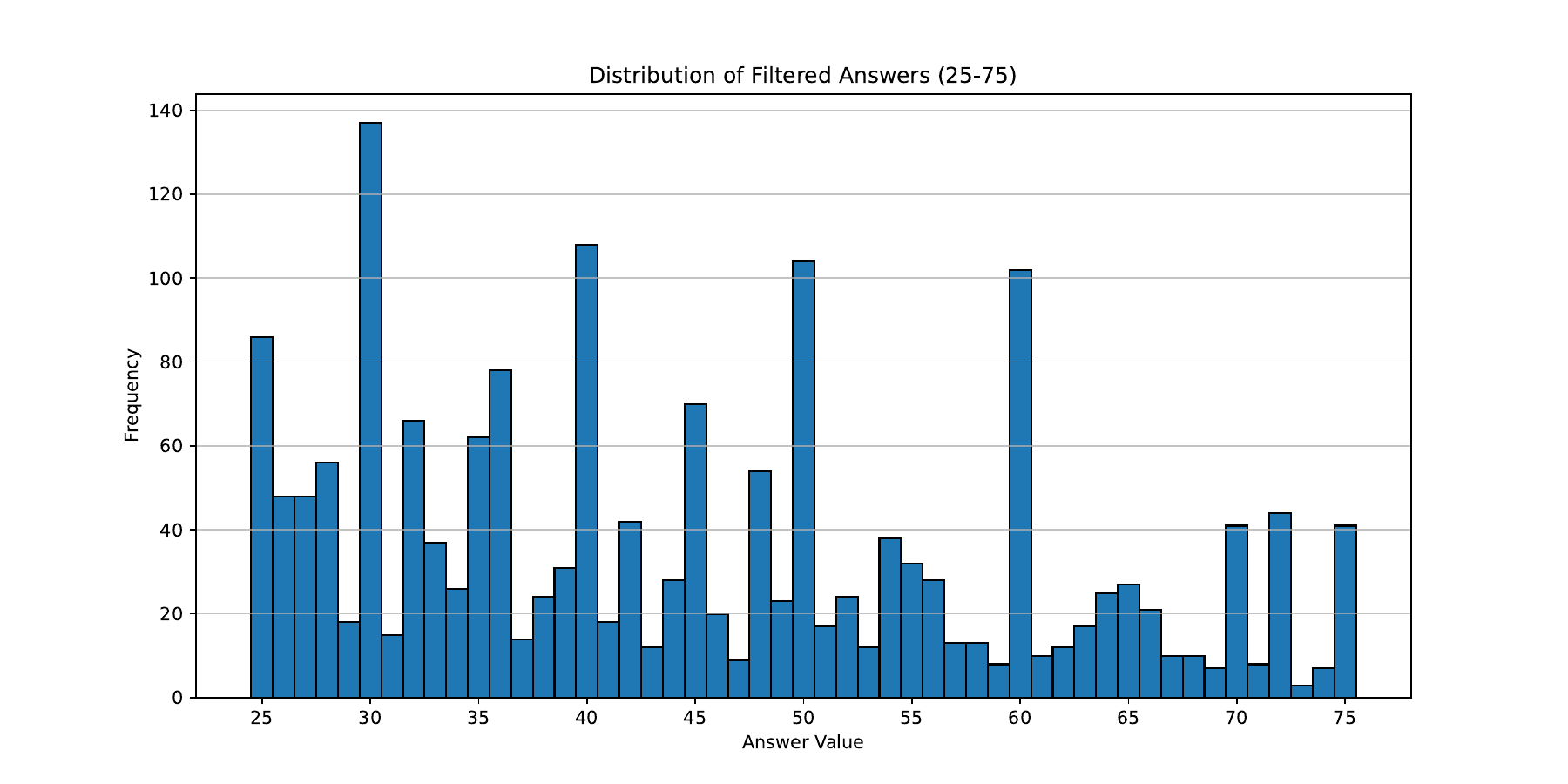}
    \caption{Distribution of the answers in GSM8K after limiting to answers in the range [25,75].}
    \label{fig:placeholder}
\end{figure}

\begin{figure}[h!]
    \centering
    \begin{subfigure}[b]{0.49\textwidth}
        \centering
        \includegraphics[width=\linewidth]{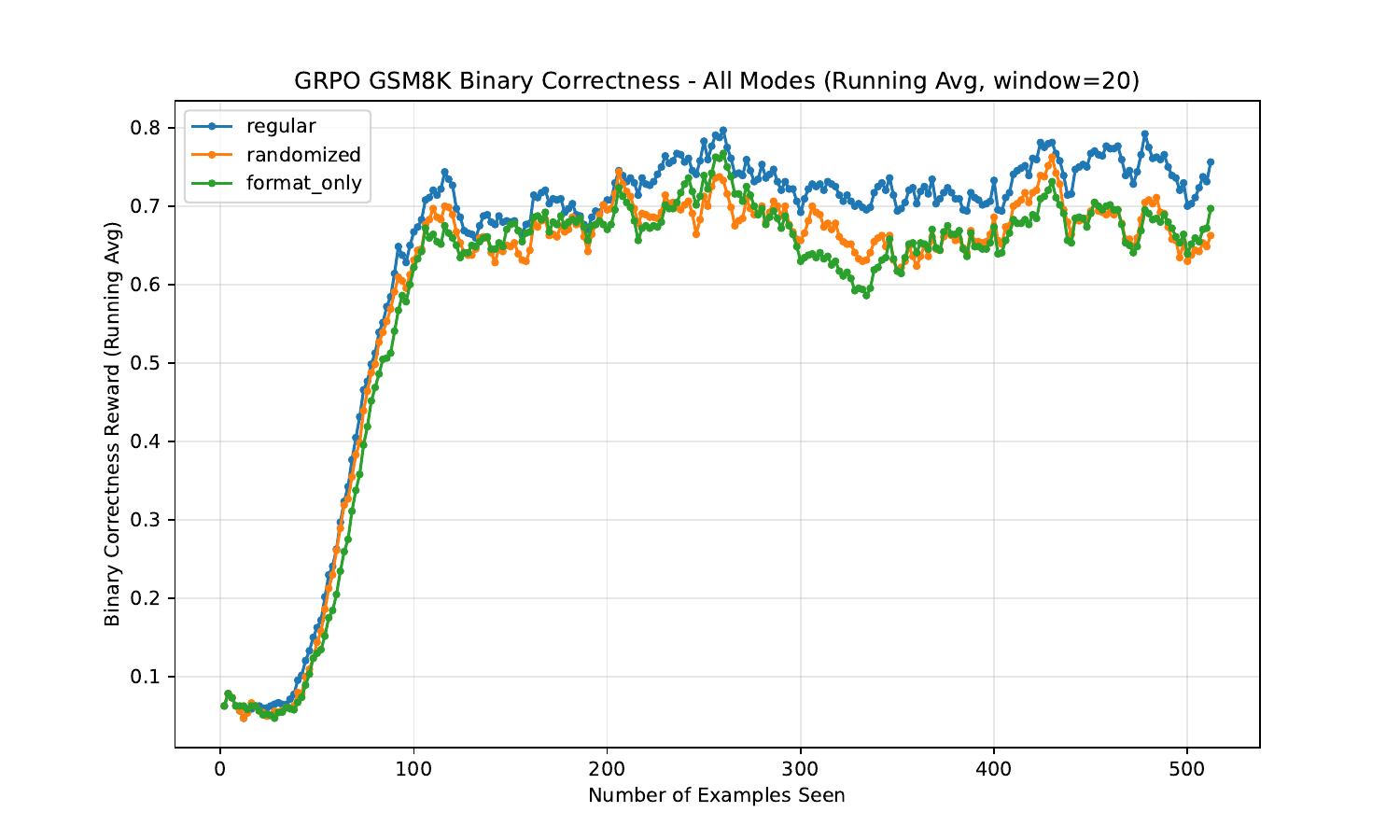}
        \caption{Run A, 512 examples.}
        \label{fig:run_a}
    \end{subfigure}
    \hfill %
    \begin{subfigure}[b]{0.49\textwidth}
        \centering
        \includegraphics[width=\linewidth]{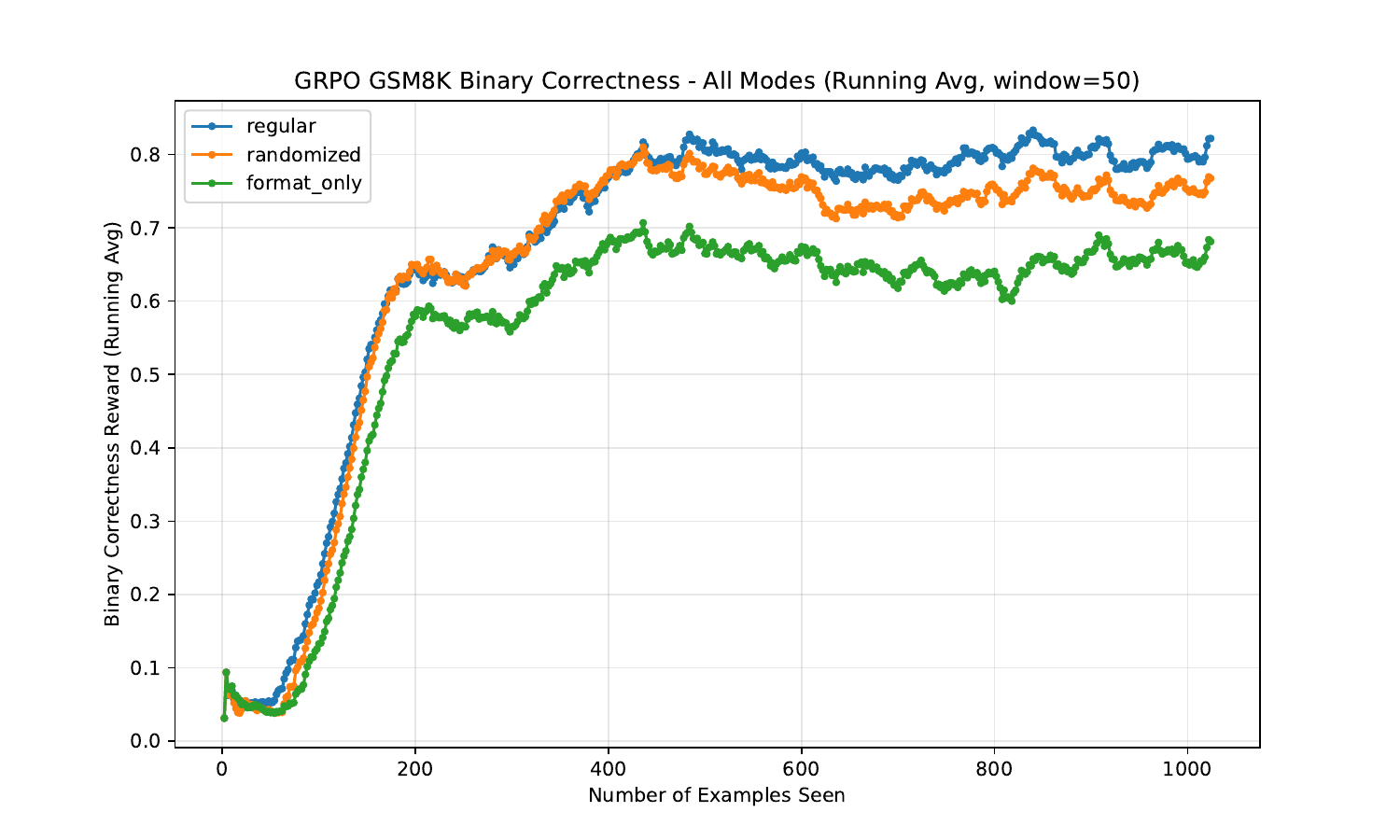}
        \caption{Run B, 1024 examples.}
        \label{fig:run_b}
    \end{subfigure}
    \caption{Average binary correctness compared to the original answer in the dataset for each of the three modes. The binary correctness is not used for training in any way (even in regular mode, a quadratic error is used instead, since we want the numbers to simulate probabilities).}
    \label{fig:combined_runs}
\end{figure}

As can be seen in Fig~\ref{fig:combined_runs}, in two different runs of the experiment, but with the bernoulli random number generation having a different seed each time, one time we see that the results of Randomized are closer to Format Only, and the other closer to Regular. This implies that the experiment is sensitive to the specific "coin flips" of our data generation for the randomized case. As such, performing a single (or two) runs are not enough to properly gauge performance. As such we proceed to perform multiple runs with different seeds and calculate a standard deviation. This will also allow us to test if increasing the number of examples while decreasing learning speed will make the model less sensitive to luck.

\begin{figure}
    \centering
    \includegraphics[width=1\linewidth]{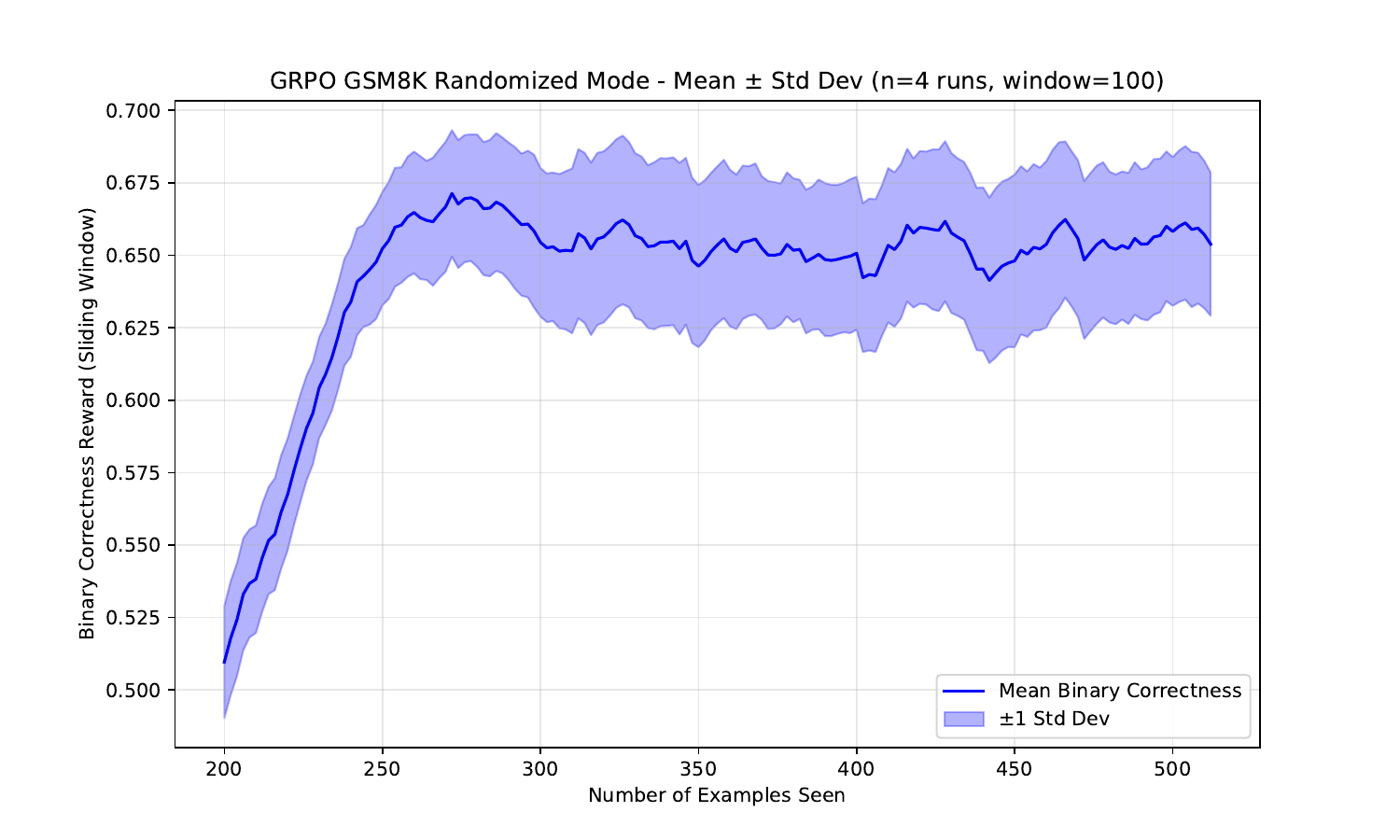}
    \caption{Window is for moving average.}
    \label{fig:4runs_aleatoric}
\end{figure}

Looking at Fig~\ref{fig:4runs_aleatoric}, we can see that the average binary correctness for the randomized mode after 512 examples is about 0.65 with a standard deviation of 0.025. And going back to Fig~\ref{fig:run_a} we can see that the performance in regular mode is 0.75 while in format only mode it is about 0.675. So in fact we can conclude that in most cases, the noise from the unreliable feedback (since it includes the result of meaningless aleatoric coin flips) is actively harmful, and it is a rare case where it leads to better than format only performance. Having the real underlying probability as the ground truth is about 4 standard deviations better.

\section{Forecasting}

Now that we have confirmed the above, we are motivated to perform our next experiment with Polymarket market probabilities as the ground truth, taken from the 512 highest volume contemporary markets. The models are provided with the question, resolution rules, and a Perplexity generated summary of relevant news.

\begin{table}[ht] \centering \begin{tabular}{lcc} \toprule Model & Start NCE & End NCE \\ \midrule Qwen 1.5B & $-1.661 \pm 0.023$ & $-0.415 \pm 0.120$ \\ Qwen 3B & $-0.975 \pm 0.035$ & $-0.305 \pm 0.007$ \\ Qwen 7B & $-0.995 \pm 0.148$ & $-0.406 \pm 0.275$ \\ \midrule Claude 3.5 Sonnet & \multicolumn{2}{c}{$-0.62 \pm 0.10$} \\ \bottomrule \end{tabular} \caption{Mean negative cross-entropy (NCE) with respect to market probabilities for the GRPO-post-trained Qwen models at the start and end of training, together with Claude 3.5 Sonnet as the reference model. Higher is better. Error terms indicate sample standard deviations.} \label{tab:qwen-cross-entropy} \end{table}

As can be seen in Table~\ref{tab:qwen-cross-entropy}, the GRPO training process improves the model's initial forecasting capability as measured by negative cross entropy with the Polymarket market prices, but only up to a limit at about $-0.48$ to $-0.3$ depending on the run. Additionally, the model size does not affect the point of plateau. In the table above there are two runs for each of the sizes 1.5B, 3B and 7B, and all reach the same point within noise (note that both the worst and best run are of Qwen2.5 7B).

As a baseline, running Claude Sonnet 3.5 on the same validation dataset, a model with a similar though later knowledge cutoff (an advantage), achieves a mean negative cross entropy of $-0.62 \pm 0.10$, below the lower end of that range. The lack of a scaling effect post training, implies that the models are being bottlenecked by the information available through the perplexity (Sonar Reasoning Pro) generated summaries. Indeed we can see that even at the end of training, all models consistently are misled when provided with incorrect information by the summarizing LLM, as seen in Example \ref{example_perplexity}.

\newtheorem{example}{Example}

\begin{example}
\label{example_perplexity}
\textbf{Prompt:}

Question: Israel x Iran ceasefire broken by July 11?

Criteria: This market will resolve to "Yes" if either Israel initiates a drone, missile, or air strike on Iranian soil or any official Iranian embassy or consulate, or Iran initiates a drone, missile, or air strike on Israeli soil or any official Israeli embassy or consulate, by 11:59 PM ET on the listed date. Otherwise, this market will resolve to "No". For the purposes of this market, a qualifying "strike" is defined as the use of aerial bombs, drones or missiles (including cruise or ballistic missiles) launched by Israeli or Iranian military forces that impact the other party’s ground territory or any official embassy or consulate (e.g., if a weapons depot on Iranian soil is hit by an Israeli missile, or an Israeli air base is hit by an Iranian drone, this market will resolve to "Yes") that is officially acknowledged by the relevant government or a consensus of credible reporting. Missiles or drones which are intercepted and surface-to-air missile strikes will not be sufficient for a "Yes" resolution regardless of whether they land on the opposing party’s territory or cause damage. Actions such as artillery fire, small arms fire, FPV or ATGM strikes directly, ground incursions, naval shelling, cyberattacks, or other operations conducted by ground operatives will not qualify. Actions taken by proxy groups (e.g. Hezbollah, Houthis, etc.) will not be considered. The resolution source will be a consensus of credible reporting. LLM 

Generated Summary: Summary generated at 2025-10-06[...] Based on the search results, this prediction market question has already been resolved, as the current date is October 6, 2025, and the market was asking about events by July 11, 2025. Key Developments The Israel-Iran ceasefire was agreed to on June 24, 2025, following their 12-day war that began on June 13, 2025[1]. However, the ceasefire appears to have been violated almost immediately. On the very day the ceasefire was announced, both sides continued military exchanges - Israel resumed strikes after agreeing to the ceasefire and accused Iran of violating the peace deal by firing missiles at Israeli cities, while Iran launched approximately 20 missiles at Israel targeting the north, south, and center of the country[1]. The search results indicate that Israeli officials reported another wave of missiles were launched from Iran after the ceasefire went into effect, with the IDF stating that two missiles fired by Iran at 10:30 a.m. local time were intercepted[1].  Market Resolution Context Given that the ceasefire was established on June 24, 2025, and there were documented violations occurring on that same day involving missile strikes by both parties, the prediction market would have resolved to "Yes" well before the July 11 deadline. The resolution criteria specified that drone, missile, or air strikes by either party on the other\'s territory would qualify for a "Yes" resolution, and the continued exchanges on June 24 clearly met these criteria[1]. The fact that both Israel and Iran accused each other of ceasefire violations and continued military strikes immediately after the agreement indicates the ceasefire was effectively broken within hours of its announcement. Provide a probability forecast (0-100) for the given question. Respond in the following format: <reasoning>...</reasoning><answer> (where 0 <= p <= 100)</answer>'

\textbf{Response:}

<reasoning>
...
</reasoning>

<answer>
100
</answer> 

\textbf{Extracted: 100 }

\textbf{Market Prediction: 0.00 }

\end{example}

\chapter{\label{ch:5-discussion}Discussion}
Between the main experiment results and the additional results, we have seen that RL can significantly increase the performance of LLMs in judgmental forecasting, and is feasible to perform on a single H100 in 2-3 hours for a $\leq$ 7B model, or a 4xH100 node and 12 hours for a 14B model. This is without quantization or LoRA.

Wikipedia revisions are a potentially powerful tool for providing a model with access to temporally restricted information, but will likely need additional research to be successful beyond simply providing a context summary.

It is likely possible to learn to train LLMs to predict Bernoulli probabilities both by seeing the ground truth probability and by seeing sampled outcomes, but having the ground truth probabilities allows orders of magnitude faster training. Using the market probabilities from a prediction market as if they are the ground truth probabilities seems to work well in practice for training. 

Out of the box, larger models are better at forecasting than smaller models. The gap can be reversed in the limited information regime through GRPO finetuning of the smaller models, but likely will require more compute, or a smaller gap in size between models, to be matched and reversed when the LLMs have access to significant amounts of information in the context. The lack of a scaling effect for the small context regime seems to imply that when the model is presented with limited information, the information available acts as a bottleneck on performance. Future experiments could consider trying to see how prediction accuracy scales with information available, or a relationship between model size and the optimal amount of context.

\startappendices
\appendix \chapter{Supplementary Material} \label{app:supplementary}

\section{Prompts Used}

Prompt used when training and evaluating models that need to provide a prediction for a forecasting question:
\begin{quote}
    
Provide a probability forecast (0-100) for the given question.

Respond in the following format:

<reasoning>
...
</reasoning>
<answer>
p (where 0 <= p <= 100)
</answer>

\end{quote}

In the GSM8K setting:

\begin{quote}
\{GSM8K question\}

Respond in the following format:
<reasoning>
...
</reasoning>
<answer>
...
</answer>
\end{quote}

In the Low Information Setting, this was the prompt provided to the Perplexity models when generating the extra context for the dataset:

\begin{quote}
 Provide a concise summary of information relevant to this Prediction Market question:

Question: \{question\}

Resolution Criteria: \{resolution\_criteria\}

\end{quote}

In the High Information Setting, this was the prompt provided to the Perplexity models when generating the extra context for the dataset:

\begin{quote}
 Provide relevant information about this question:

Question: \{question\}

Resolution Criteria: \{resolution\_criteria\}

Do not provide a forecast, only relevant information for a forecaster to consider.   
\end{quote}

\section{Experiment Details}
\label{section:experiment_details}
\subsection{Hyperparameters}
During GRPO training, gradient accumulation was always 4, the number of samples generated per question was 16 (or 32 in the high information setting). The choice of number of samples generated (between 16 and 32) did not seem to significantly affect results. Batch size was always 8 per GPU. The maximum prompt length was 512 tokens in the low information setting and 4096 in the high information setting. The maximum completion length was 1024 and 2048 tokens respectively. When negative Cross Entropy or KL Divergence were used as reward functions, reward was capped to be at least -5.0 for both the models under training and evaluation (including for the models used to benchmark, for fairness), to prevent infinities.

The learning rate was decided on using a search of all learning rates in range 1e-6 and 1e-5, and the learning rate schedule was Cosine.

\subsection{Models Used}
The main GRPO experiments were all performed on the Qwen2.5 Instruct series of models, with sizes 1.5B, 3B and 7B being the main ones used. For the Wikipedia tool calling experiment described in \ref{wikipedia_rev}, also the 14B size was used. As a baseline, Claude Sonnet 3.5 was used. It was chosen as it is one of the best models with a knowledge cutoff before 2025.

For summarized contemporary context generation, Sonar Reasoning Pro was used in the limited information setting and both Sonar and Sonar Reasoning Pro were used in the high information setting.

\subsection{Dataset Creation}
When the ground truth was outcome, questions collected from Manifold and Metaculus were used. When the ground truth was a probability, Polymarket questions were used. In the case of Manifold and Metaculus, questions were filtered to remove the low quality questions from the site similarly to the process described by \cite{halawi2024approaching}. In the case of Polymarket, only the 512 highest volume questions were used. In both cases the official APIs of the websites were used. For Wikipedia revisions, the official Wikipedia API was used to get access to revisions.

\subsection{Wikipedia Revisions Tool}
The Wikipedia tool allowed models to provide a timestamp and an article title, as well as an initial and ending character index (e.g. 0:1000), and the tool would provide the LLM with the requested characters of the newest revision of the closest matching article from before the forecasting question's timestamp. The LLM could use up to 5 tool calls. We also attempted to provide the LLM with an option to instead ask a stronger LLM, Qwen 72B, to read the entire article and answer a specific question for the LLM during training, though this did not affect results but was slow and expensive.

\subsection{Compute Resources}
The majority of the large experiments were performed on Oxford University's ARC cluster, on a 4xH100 node or on a single H100 GPU. The smaller experiments were performed on Vast.AI, using RTX 3090 and 4090 nodes. All experiments can be run on their respective compute in under 12 hours, and as such all experiments described in this thesis can be run on a 4xH100 node in under 12 hours.

\section{Additional Results}
\label{appendix_additional_results}
\subsection{Pretrained LLM Calibration Graph}
\begin{figure}[h!]
    \centering
    \includegraphics[width=1\linewidth]{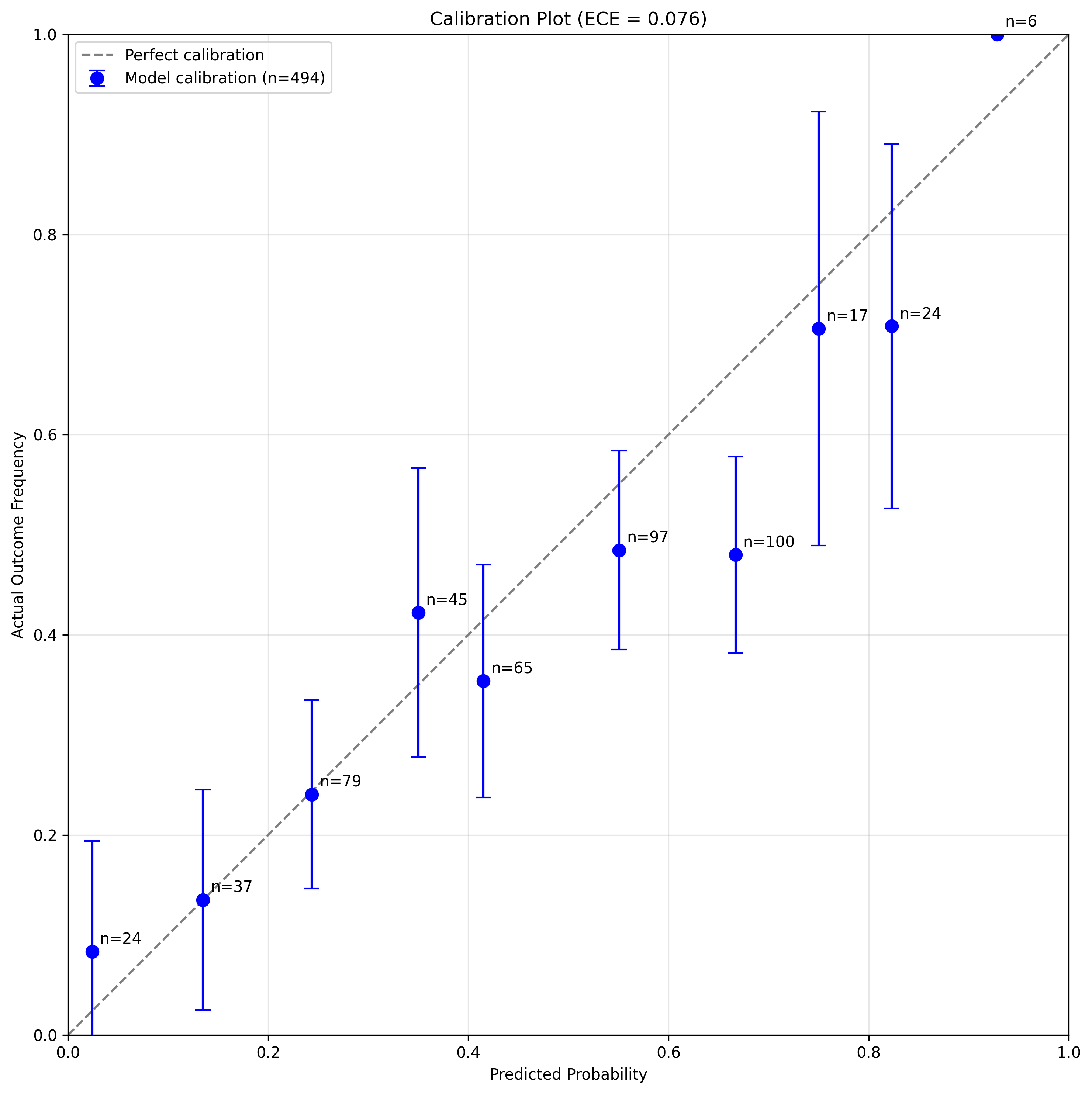}
    \caption{These are 95\% confidence intervals - just one of the dots, the one for 60-70\% probability, is clearly outside - with things happening only like 50\% of the time if predicted as 60-70. Which is interesting. It is also the largest group - a full 20\% of the model's predictions are placed there.}
    \label{fig:r1_calibration}
\end{figure}

For the DeepSeek R1 calibration graph, see Figure~\ref{fig:r1_calibration}. It provides one reason why GRPO finetuning can easily improve performance.

\subsection{High Information Setting}
In the high information setting, it seems that while the small, finetuned models improved somewhat, the large pretrained model we use as a comparison, Claude Sonnet 3.5, improved significantly more - likely because the extra information let the model's extra fundamental capability more room to shine.

\subsection{Genetic Algorithm for Prompt Refinement}
One of the directions that did not succeed in the project was using a learned prompt to guide the LLM forecasting, instead of changing the weights of the model through finetuning (RL or otherwise). This was found to barely affect the model's performance, regardless of prompt used, similarly to what was found by \cite{halawi2024approaching}.

The method used to optimize the prompts was EvolutionaryGPT \cite{Levy_EvolutionaryGPT}.

\subsection{Wikipedia Revision Tool Calls}
\label{wikipedia_rev}
 Another failed direction was using already resolved forecasting questions for training, using the true resolution as the ground truth and using time-restricted Wikipedia revision tool calls to get information for the LLM without temporal leakage. Unfortunately the LLMs consistently stabilized with the behaviour of outputting $50\%$ odds for all questions (Brier score 0.25). This is likely because training based on final outcomes only is too noisy to allow for efficient training at this scale.

The GRPO + tool call training pipeline was implemented using the Verifiers framework \cite{brown_verifiers_2025}.

\section{Note Added in Version 2}
\label{sec:v2-note}

This note was added on 25 October 2026. The rest of the document is unchanged from version 1.

\paragraph{Example~\ref{example_perplexity}.}
Some readers have interpreted Example~\ref{example_perplexity} as temporal
leakage, since the summary was generated after the market's deadline had
passed. This is not temporal leakage. In the Polymarket setting, the time of
forecast is the time at which the summary and the market price were collected,
so all information up to that point is available both to the model and to the
market whose price serves as the ground truth; at that time, this market was
trading at $0.006$. The example was chosen because the summary misstates what
happened, which leads the model to fail on an otherwise trivial question.

\paragraph{Market prices in summaries.}
Of the 1{,}024 summaries (512 markets across the two information regimes), one
contained the exact Polymarket price at the time of forecast, and seven
contained figures close to it. A further ten contained stale prices that
differed substantially from the price at the time of forecast. We treat the
eight exact or close cases ($0.8\%$), four in each information regime, as
genuine leakage. Since the per-question reward is capped at $-5$, the four in
the low-information regime reported in Table~\ref{tab:qwen-cross-entropy} can
change the gap in mean negative cross entropy between any two models by at
most $5 \times 4/512 \approx 0.039$. This is well below the gap of
$0.2$ to $0.3$ between the fine-tuned models and Claude 3.5 Sonnet. Moreover,
the same summaries were provided to Claude 3.5 Sonnet, so the leakage does not
systematically favour the fine-tuned models.

\paragraph{Additional limitations.}
Training used a single epoch, as such train reward and validation reward are the same. The last 128 questions seen in the epoch were treated as the validation set. As such, at the time the model generated its validation forecasts, it had never seen any of those questions before. Looking back at this research, I regret that I did not temporally separate validation and training, as I do now, since it is good practice, but I will comment that in practice my current experience is that this does not affect results. Additionally, some of the experiments, such as the GSM8K and Wikipedia ones, were performed before \cite{bereket2025uncalibratedreasoninggrpoinduces} showed removing the standard deviation in GRPO improves calibration. Even for the ones performed after, I was not yet aware of the results and so did not change the reward. It is a regret of mine that I did not notice this myself nor apply this result to the outcome-based training experiments, and I believe this weakens the outcome-based results in this thesis.

\setlength{\baselineskip}{0pt} %

{\renewcommand*\MakeUppercase[1]{#1}%
\printbibliography[heading=bibintoc,title={\bibtitle}]}

\end{document}